\documentclass[conference]{IEEEtran}
\IEEEoverridecommandlockouts

\usepackage{balance}
\usepackage{multirow}
\usepackage{caption}
\usepackage{subcaption}
\usepackage{graphicx}
\usepackage{array}
\usepackage{tabularx}
\usepackage{multirow}
\newcolumntype{L}[1]{>{\raggedright\arraybackslash}m{#1}}
\newcolumntype{C}[1]{>{\centering\arraybackslash}m{#1}}
\newcolumntype{R}[1]{>{\raggedleft\arraybackslash}m{#1}}
\usepackage{balance}
\usepackage{cite}
\usepackage{amsmath,amssymb,amsfonts}
\usepackage{pgfplots}
\pgfplotsset{compat=newest}
\usepackage{algorithm,algpseudocode}
\usepackage{textcomp}
\usepackage{adjustbox}
\usepackage{xcolor}
\usepackage{tikz}
\usepackage{booktabs}
\usepackage{tabularx} 

\def\BibTeX{{\rm B\kern-.05em{\sc i\kern-.025em b}\kern-.08em
    T\kern-.1667em\lower.7ex\hbox{E}\kern-.125emX}}

\begin{document}

\title{A Robust Federated Learning Approach for Combating Attacks Against IoT Systems \\Under non-IID Challenges}
\author{
    \IEEEauthorblockN{
        Eyad Gad\IEEEauthorrefmark{1}$^1$, 
        Zubair Md Fadlullah\IEEEauthorrefmark{1}$^2$, 
        and Mostafa M. Fouda\IEEEauthorrefmark{3}\IEEEauthorrefmark{4}$^3$
    }
    \IEEEauthorblockA{\IEEEauthorrefmark{1}Department of Computer Science, University of Western Ontario, London, ON, Canada. \\
        }
    \IEEEauthorblockA{\IEEEauthorrefmark{3}Department of Electrical and Computer Engineering, Idaho State University, Pocatello, ID, USA. \\
    }
    \IEEEauthorblockA{\IEEEauthorrefmark{4}Center for Advanced Energy Studies (CAES), Idaho Falls, ID, USA.\\}
    
Emails: $^1$egad@uwo.ca, $^2$zfadlullah@ieee.org, 
$^3$mfouda@ieee.org
    }

\maketitle
\IEEEoverridecommandlockouts
\IEEEpubid{\makebox[\columnwidth]{979-8-3503-8532-8/24/\$31.00~\copyright2024 IEEE \hfill} \hspace{\columnsep}\makebox[\columnwidth]{ }}
\maketitle
\IEEEpubidadjcol 

\begin{abstract}
In the context of the growing proliferation of user devices and the concurrent surge in data volumes, the complexities arising from the substantial increase in data have posed formidable challenges to conventional machine learning model training. Particularly, this is evident within resource-constrained and security-sensitive environments such as those encountered in networks associated with the Internet of Things (IoT). Federated Learning has emerged as a promising remedy to these challenges by decentralizing model training to edge devices or parties, effectively addressing privacy concerns and resource limitations. Nevertheless, the presence of statistical heterogeneity in non-Independently and Identically Distributed (non-IID) data across different parties poses a significant hurdle to the effectiveness of FL. Many FL approaches have been proposed to enhance learning effectiveness under statistical heterogeneity. However, prior studies have uncovered a gap in the existing research landscape, particularly in the absence of a comprehensive comparison between federated methods addressing statistical heterogeneity in detecting IoT attacks. In this research endeavor, we delve into the exploration of FL algorithms, specifically FedAvg, FedProx, and Scaffold, under different data distributions. Our focus is on achieving a comprehensive understanding of and addressing the challenges posed by statistical heterogeneity. In this study, We classify large-scale IoT attacks by utilizing the CICIoT2023 dataset. Through meticulous analysis and experimentation, our objective is to illuminate the performance nuances of these FL methods, providing valuable insights for researchers and practitioners in the domain.
\end{abstract}

\begin{IEEEkeywords}

 Internet of Things (IoT), Federated Learning (FL), non-IID data, statistical heterogeneity, Intrusion Detection System (IDS), scaffolding technique. 

\end{IEEEkeywords}

\section{Introduction}
In the landscape of Internet of Things (IoT) networks, generating extensive datasets has become a crucial component of contemporary technological progress. The significance of data as a novel production medium is underscored by the vast possibilities enabled through data sharing and mining~\cite{10379499, agrawal2000privacy,
hynes2018demonstration,niu2017trading}. Notably, the potential benefits are evident in scenarios such as the sharing of vehicle IoT data to alleviate congestion and the collaborative analysis of medical big data to facilitate disease prediction~\cite{gad2023novel, gad2023advancing}. However, the conventional methods of data sharing, which involve direct uploads to centralized repositories, prove insufficient in meeting the escalating demands for robust data security and privacy protection. This inadequacy becomes particularly pronounced when considering the susceptibility of the network traffic and transmitted data across IoT devices to various cyber-attacks.
In securing IoT networks against evolving these attacks, traditional machine learning (ML) has been utilized in Intrusion Detection Systems (IDS) to maintain network integrity. However, the computational limitations of IoT devices pose challenges for implementing effective ML-based IDS. The constrained computing capabilities of IoT devices make it difficult to process the substantial data volumes required for IDS model training. Additionally, privacy concerns arise with centralized learning approaches in IoT networks, as they involve collecting data or network traffic, potentially compromising sensitive information~\cite{10278423,10454870,10000881}. Balancing privacy and effective intrusion detection are essential for establishing trust in IoT systems. To address these challenges, Federated learning (FL) is proposed by decentralizing model training to edge devices or parties, providing an innovative solution that enables collaborative model training without exposing raw data to a central server. FL has attracted many research
interests~\cite{dai2020federated,deng2020distributionally,eustache2017attribution,dinh2020personalized,felix2020federated,guha2019one,hasan2020deep, gad2022federated,9764093} and been widely
used in practice~\cite{he2020group,hard2018federated,hanzely2020lower,NASSER2022108672}. This enhances privacy and security and leverages collective network intelligence for improved intrusion detection accuracy and efficiency~\cite{nguyen2019diot, 9606815, gad2023communication2, gad2023communication, 10437358}. 

One common challenge in FL arises from statistical heterogeneity~\cite{li2021evaluating,tang2021data, 9975908, 10437327}, which encompasses variations in data distribution and characteristics among local datasets, including differences in class balance or data quality, commonly referred to as non Independently and Identically Distributed (non-IID). Yet, addressing this heterogeneity proves especially challenging in securing IoT networks. The varied distributions of IoT attacks~\cite{fredrikson2015model} underscore the significance of addressing this heterogeneity, as its presence in FL significantly impacts model accuracy, convergence speed, and overall performance.

However, Existing literature in IoT attack detection reveals a gap in comprehensive comparisons between FL methods, especially those designed to address statistical heterogeneity. This gap contributes to the limited understanding of their comparative performance. The need for such a comparison arises from the unique challenges posed by statistical heterogeneity in diverse local datasets, a critical consideration in the field of IoT security. To fill this gap, our objectives and contributions are to conduct a thorough examination of FL algorithms concerning the classification of IoT attacks. This involves analyzing and comparing FedAvg~\cite{mcmahan2023communicationefficient}, FedProx~\cite{li2018}, and Scaffold~\cite{karimireddy2021scaffold}, which are known for addressing statistical heterogeneity challenges. Additionally, our study examines the CICIoT2023~\cite{ciciot23} dataset under both Independently and Identically Distributed (IID) and non-IID data distributions. The dataset is recently published and specifically tailored for large-scale attacks in IoT environments, which has not been previously utilized in existing literature on FL in this domain.

The rest of the paper is organized as follows. In Section~\ref{sec:background}, we provide a comprehensive review of related work, surveying existing methods and approaches for addressing non-IID. Our proposed study is detailed in Section~\ref{sec:methods}, which is divided into other subsections that introduce the FL framework, dataset, FL methods, and data partitioning. Section~\ref{sec:exp} outlines our FL experimental setup, IID results, and non-IID results. Finally, Section~\ref{sec:conclusion} summarizes and discusses our findings. 

\section{Related Work}\label{sec:background}

Many studies~\cite{hsu2019measuring,hu2020oarf,huang2005maximum,hynes2018demonstration,kairouz2019advances} have provided substantial evidence supporting the efficacy of FL across various domains. Notably, in the realm of IDS, FL has demonstrated favorable outcomes. For instance, Nguyen Duc Thien \textit{et al.}~\cite{nguyen2019diot} proposed an FL-based IDS model strategically positioned at the security gateways of participating network computers, aimed at automatically detecting threats in IoT devices.

Similarly, Nguyen Chi Vy \textit{et al.}~\cite{vy2021federated} devised a sophisticated deep learning model combined with FL to construct an IDS specifically designed to defend against poisoning attacks within the Industrial IIoT. The collaboration between local devices in the FL framework enables the construction of a robust and adaptive IDS that leverages collective intelligence without compromising data privacy.

Anastasakis \textit{et al.}~\cite{anastasakis2022} suggested an innovative approach by incorporating Differential Privacy into FL for IoT systems, striking a balance between accurate attack detection and safeguarding data privacy. While these studies are commendable, they often lack a nuanced client weighting system and a robust class balancing technique, both of which are crucial components introduced in the work presented in this paper.

Additionally, Beibei Li \textit{et al.} introduced DeepFed~\cite{li2021deepfed}, a comprehensive FL scheme tailored for IDS in industrial cyber-physical systems. DeepFed employs collaborative training of a deep learning model on security agents, utilizing a combination of Convolutional Neural Networks and Gated Recurrent Units for enhanced accuracy and adaptability. Attota \textit{et al.}~\cite{attota2021} presented MV-FLID, a pioneering multi-view ensemble learning approach that significantly enhances attack detection accuracy across various attack classes. This approach considers multiple perspectives or views of the data to improve the overall robustness of the FL-based IDS.
Moreover, Anastasakis \textit{et al.}~\cite{abhijit2023} underscored the significant potential of FL in bolstering IoT cybersecurity, emphasizing its role in enhancing privacy and reducing data transfer costs. These studies collectively contribute to the application of FL in the domain of IDS.

\section{System Model and Proposed Method}
\label{sec:methods}
\subsection{Proposed FL System Model}
In our proposed FL framework, 
the network architecture is organized into three main sections. The foundational section comprises IoT or user devices, each connected to a designated base station \(k\). Each base station is represented as a client within the FL framework, possessing its unique local data repository and IDS. These base stations act as training participants, utilizing their localized network traffic data from IoT devices to train the FL model.

The FL model begins with the participant clients training on their local data. Following the training phase, these participant clients generate a model update representing the changes made to their local model during training. Subsequently, they transmit their model parameters $w_k$, referred to as model updates, to the central network infrastructure or cloud which is represented as the FL server or global model. The server aggregates the model updates from all participant clients and redistributes the refined aggregated model parameters $w_o$ back to each client, referred to as model download, initiating the subsequent round of training. 

\subsection{Data Sources and Preparation}

\begin{figure}[!htb]
\centering
\includegraphics[width=\linewidth]{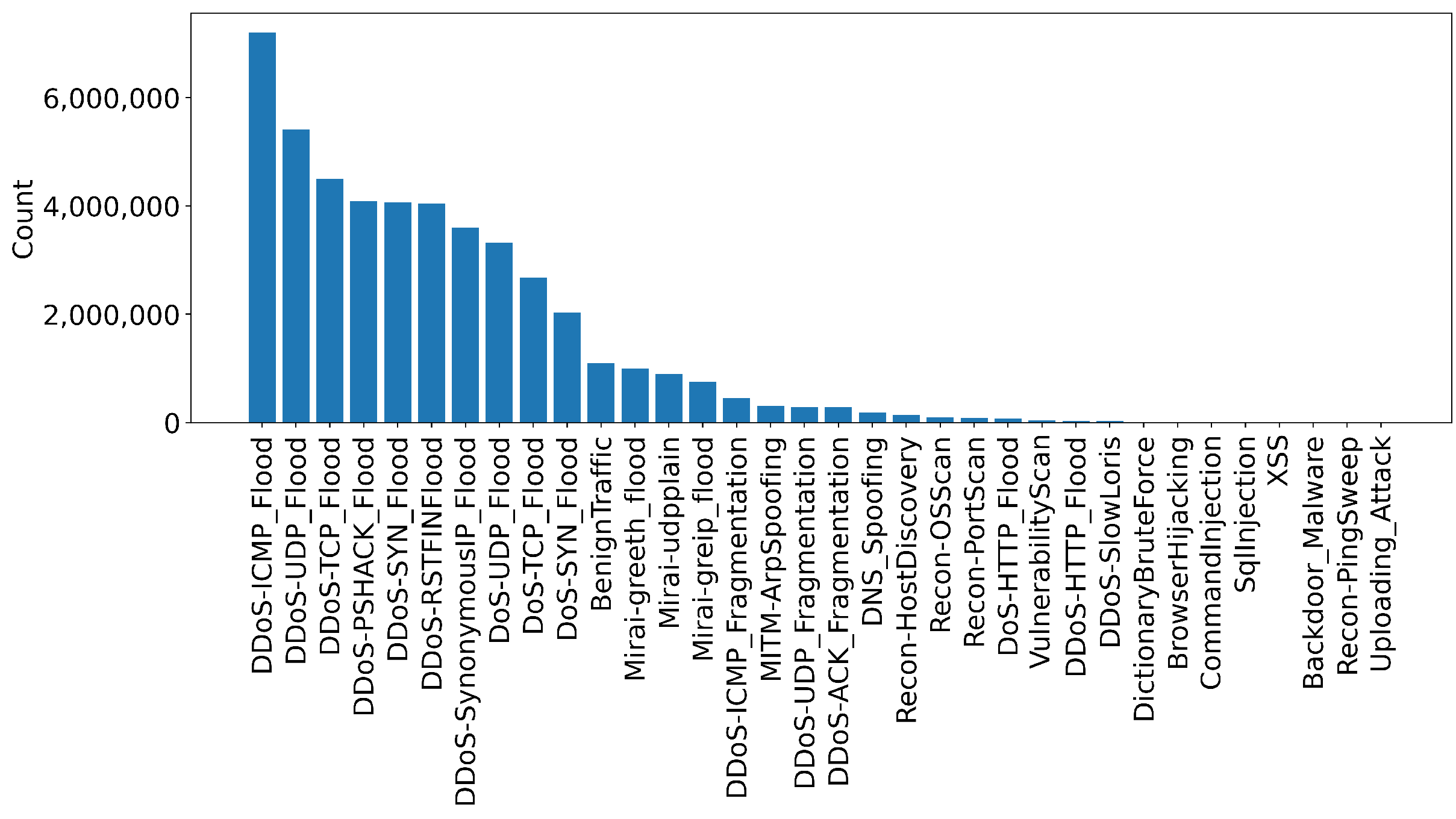}
\caption{Number of data points for each attack.}
\label{fig:attacks33}
\end{figure}

In our work, the CICIoT2023 dataset is employed as the foundational data source. The CICIoT2023 dataset serves as a dynamic platform tailored for the exploration of expansive cyber-attacks within the IoT environments. This dataset is constructed upon an intricate network architecture that encompasses over 105 distinct IoT devices. For each enumerated attack strategy, the dataset is meticulously curated to incorporate benign packet transmissions. This inclusion ensures the emulation of authentic network traffic scenarios through packet captures, thereby facilitating a more nuanced analysis.

\begin{figure}[!t]
\centering
\includegraphics[width=\linewidth]{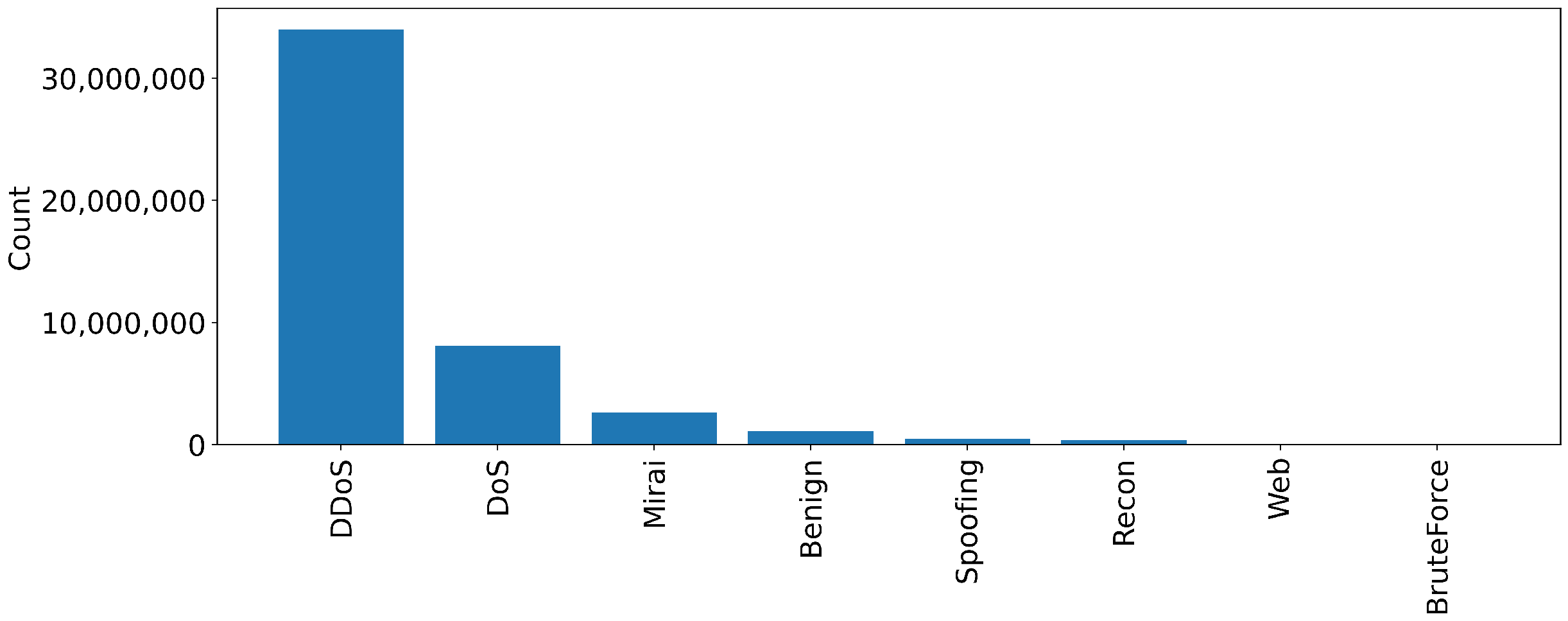}
\caption{Number of data points for each attack.}
\label{fig:attacks8}
\end{figure}

As illustrated in Fig.~\ref{fig:attacks33}, the dataset encapsulates 33 attacks, which can be systematically categorized into seven categories. These classifications encapsulate DDoS, DoS, Reconnaissance, Web-based assaults, Brute Force techniques, Spoofing endeavors, and Mirai-based attacks including the benign category, as in Fig.~\ref{fig:attacks8}. Notably, certain attack modalities exhibit characteristics that render them pertinent to multiple categories. A case in point is the TCP Flood and UDP Flood, both of which have been categorized under both DDoS and DoS classifications. The dataset encompasses a comprehensive array of 46 distinct features. These features span a multitude of parameters, including but not limited to, flow duration, header length, and protocol specifications such as HTTP, HTTPS, and DNS. Collectively, this extensive feature set provides an enriched perspective on network behavioral patterns.

In the initial phase of data preparation for FL, the benign category is systematically distributed across the remaining seven categories. Subsequently, for the server data utilized in testing the global model,  20\% of the data is extracted from each of these categories.

Moreover, To prevent any impact of the challenges associated with data imbalance, a portion is sampled from each category, considering the number of data points for each attack. This approach results in achieving a balanced attacks within each category, which enhances the effectiveness of classification algorithms, preventing potential biases towards the over represented majority class that might otherwise occur. 

\subsection{IID and non-IID Partitioning}
\begin{figure}[!t]
  \centering
  \begin{subfigure}{0.9\linewidth}
  \centering
  \includegraphics[width=0.9\linewidth]{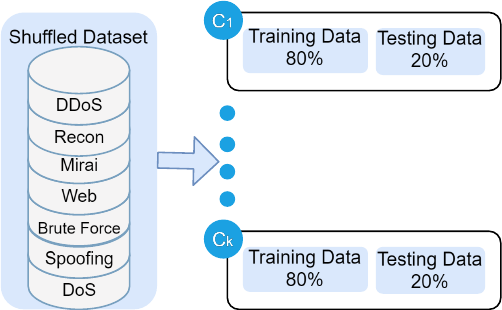}
    \caption{IID partitioning.}
    \label{fig:iid-clientsdata}
  \end{subfigure}
  \begin{subfigure}{0.9\linewidth}
  \centering
  \includegraphics[width=0.9\linewidth]{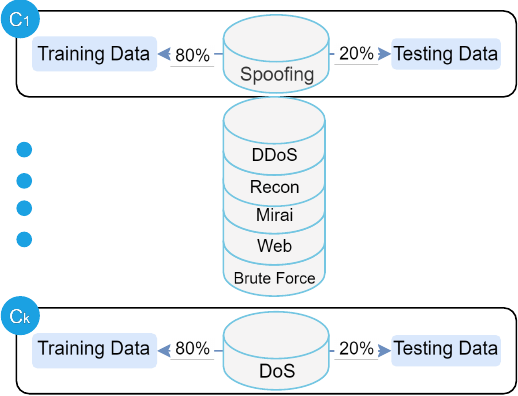}
    \caption{non-IID partitioning.}
    \label{fig:noniid-clientsdata}
  \end{subfigure}
  \caption{Clients' local data preparation.}
  \label{fig:clientsdata}
\end{figure}

Regarding client data partitioning, the FL framework adopts two partitioning approaches. In the first approach, referred to as the IID partitioning approach as depicted in Fig.~\ref{fig:iid-clientsdata}, data from all seven categories undergoes a random shuffle before being distributed among a predetermined number of clients. In IID, participant clients receive an equal distribution of classes, ensuring they possess the same characteristics.
Conversely, the second approach, referred to as the non-IID partitioning approach as depicted in Fig.~\ref{fig:noniid-clientsdata}, designates each category as an individual client. With seven distinct categories, this results in seven clients, each encapsulating its unique classes and attack characteristics.
\subsection{FL methods}
In this section, we delve into three prominent FL methods: FedAvg, FedProx, and Scaffold.

FedAvg is a foundational FL algorithm that forms the basis for subsequent advancements. It operates through a cyclic process of local model training and global model aggregation. Let \( w_t^i \) represent the local model of client \( i \) at round \( t \), and \( P_i \) denote the set of participating clients. The global model \( w_{t+1} \) is updated by averaging the local models weighted by the number of samples each client has. The update formula for FedAvg is as follows:
\begin{equation}
    w_{t+1} = w_t - \eta \frac{1}{|P_t|} \sum_{i \in P_t} \frac{|D_i|}{n} w_t^i
    \label{eq:fedavg}
\end{equation}
Here, \( \eta \) is the learning rate, \( |P_i| \) represents the number of participating clients, and \( |D_i| \) denotes the local dataset size of client \( i \).

FedProx enhances FedAvg by incorporating an additional L2 regularization term in the local objective function to control the size of local updates. The objective function for FedProx is given by:
\begin{equation}
    L(w; b) = \frac{1}{|P_t|} \sum_{i \in P_t} |D_i| \ell(w; b) + \frac{\mu}{2} \|w - w_t^i\|^2
    \label{eq:fedprox}
\end{equation}
Here, \( \ell(w; b) \) is the standard loss function, \( \mu \) is the hyper-parameter controlling the regularization strength, and \( \|w - w_t^i\| \) is the L2 norm. The proximal term enforces the model parameters to be close to the previous iteration, while the regularization term promotes sparsity in the model

Finally, Scaffold addresses non-IID challenges by introducing variance among the parties and applying variance reduction techniques. The algorithm uses control variates for both the server (\( c \)) and parties (\( c_i \)) to estimate the update directions. The update formula for Scaffold is:
\begin{equation}
    w_{t+1} = w_t - \eta \frac{1}{|P_t|} \sum_{i \in P_t} \frac{\tau_i}{\sum_{i \in P_t} \tau_i} \Delta w_t^i
    \label{eq:scaffold}
\end{equation}
Here, \( \tau_i \) represents the control variate of client \( i \), and \( \Delta w_t^i \) is the corrected local update accounting for the drift in local training.

\section{Experimental Results}\label{sec:exp}
In this section, we present our empirical results to validate our proposed FL framework. First, we provide the data modeling outcomes. Then, we describe the experimental setup for the FL framework. Next, we present the results involving IID data. Finally, the non-IID experimental results and discussions are provided. 
\subsection{Data Modeling}
Before embarking on the FL implementation, we meticulously assessed the dataset using four distinct ML methodologies, namely Logistic Regression (LR), AdaBoost, Random Forest (RF), and Artificial Neural Network (ANN). These methods have demonstrated efficacy across diverse applications, including cybersecurity. 
\begin{table}[!htb]
    \centering
    \caption{Performance metrics for different classification algorithms.}
    \label{tab:classification_metrics}
    \renewcommand{\arraystretch}{1.4}
    \begin{tabular}{|l|l|c|c|c|c|}
    \hline
        \textbf{Class Type} & \textbf{Metric} & \textbf{LR} & \textbf{AdaBoost} & \textbf{RF} & \textbf{ANN} \\
        \hline
        \multirow{4}{*}{34 classes} & Accuracy & 80.23 & 60.79 & 99.16 & 98.61 \\
        & Recall & 59.52 & 60.77 & 83.16 & 73.19 \\
        & Precision & 48.68 & 47.96 & 70.45 & 66.53 \\
        & F1-score & 49.39 & 47.35 & 71.4 & 67.23 \\
        \hline
        \multirow{4}{*}{8 classes} & Accuracy & 83.17 & 35.14 & 99.44 & 99.11 \\
        & Recall & 69.61 & 48.78 & 91 & 90.66 \\
        & Precision & 51.24 & 46.49 & 70.54 & 67.94 \\
        & F1-score & 53.94 & 36.87 & 71.93 & 69.73 \\
        \hline
        \multirow{4}{*}{2 classes} & Accuracy & 98.9 & 99.59 & 99.68 & 99.44 \\
        & Recall & 89.04 & 94.73 & 96.52 & 93.33 \\
        & Precision & 86.32 & 96.56 & 96.54 & 94.76 \\
        & F1-score & 87.63 & 95.63 & 96.53 & 94.03 \\
        \hline
    \end{tabular}
\end{table}

Table~\ref{tab:classification_metrics} delineates the performance of these algorithms under three classification scenarios: binary classification (malicious vs. benign), multiclass with eight categories (benign vs. different attack types), and a more granular multiclass with 34 categories (benign vs. individual attacks). The corresponding performance metrics, encompassing accuracy, recall, precision, and F1-score, are tabulated for a comprehensive evaluation, as presented in Table~\ref{tab:classification_metrics}.

In the binary classification, all methods demonstrate strong accuracy above 98\%; for the eight-class attack categorization, LR and Adaboost exhibit reduced accuracy and notably diminished F1-scores, contrasting with the consistent performance of RF and ANN. These latter methods maintain around 70\% F1-scores. In the intricate 34-class attack classification, both RF and ANN sustain high accuracy and F1-scores, marginally decreasing by approximately 1\% from the eight-class scenario. Conversely, LR and Adaboost consistently underperform, achieving below 80\% accuracy and less than 50\% F1-scores across all cases.

\subsection{FL Experimental Setup}
In the experimental evaluation, we rigorously examined the performance and effectiveness of FedAvg, FedProx and Scaffold algorithms across various configurations, focusing particularly on a non-IID dataset that mirrors the complex data distributions encountered in real-world scenarios. The experiments were designed with specific setups, including scenarios with 5 clients under an IID setting and 7 clients, each aligned with distinct categories in the non-IID setting. For the FedProx configurations, three runs were conducted across mu values of 0.1, 0.2, and 0.4. The experiments involved 100 rounds of FL iterations, each comprising 10 epochs, and utilized a learning rate set to 0.01 a decay factor of 0.8. Additional parameters incorporated Cross-Entropy Loss as the primary loss function, and Stochastic Gradient Descent (SGD) as the optimization algorithm, aiming for optimal convergence and performance assessment.

\subsection{IID Experimental Results}
\begin{figure}[!htb]
\centering
\begin{subfigure}{\linewidth}
\includegraphics[width=\linewidth]{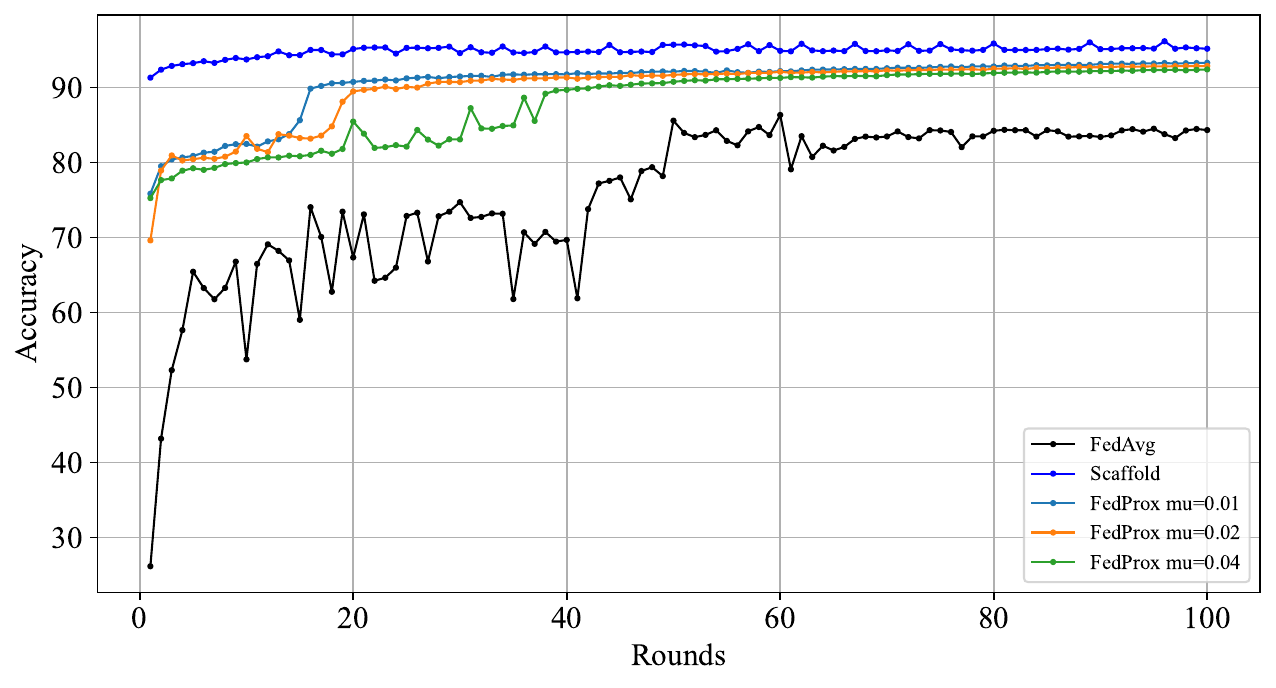}
\caption{Comparison of accuracy of the collaborative learning under IID considerations with existing methods.}
\label{fig:iid-global-acc}
\end{subfigure}
\begin{subfigure}{\linewidth}
\includegraphics[width=\linewidth]{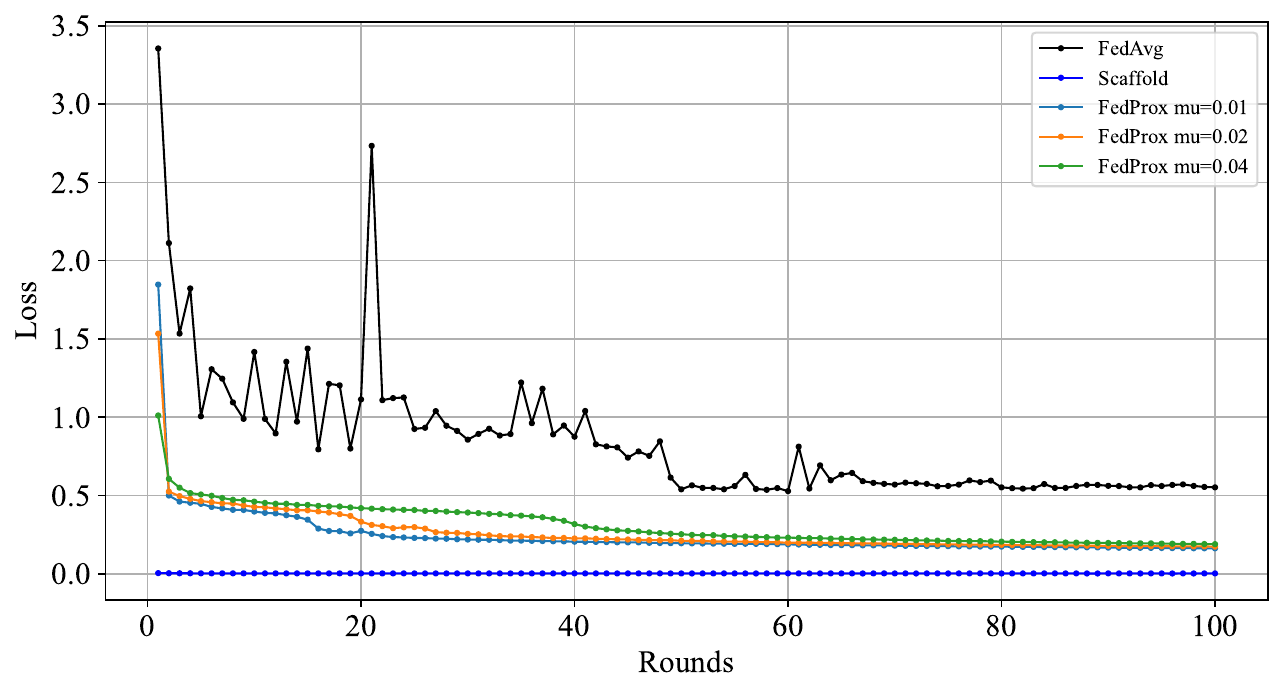}
\caption{Comparison of Loss incurred in collaborative learning under IID considerations with existing methods.}
\label{fig:iid-global-loss}
\end{subfigure}
\caption{Global performance outcomes of the proposal under IID data in contrast with conventional methods.}
\label{fig:iid-global}
\end{figure}

Fig.~\ref{fig:iid-global} illustrates the results of the global model's performance. The experimental results reveal interesting patterns across the FL algorithms evaluated under the IID setting. In the case of FedAvg, the global accuracy and loss show fluctuations over the 100 rounds, reaching a peak accuracy of 86.33\% and a minimum loss of 0.53 at the 60th round. For the FedProx algorithm with mu=0.01, there is a consistent improvement in global accuracy, peaking at 93.29\%, while the loss steadily decreases to 0.16. However, varying the mu to 0.02 or 0.04 shows a slight decrease in accuracy and an increase in loss. Notably, Scaffold consistently achieves remarkable performance with constantly high accuracy of up to 96.16\%, indicating stable and superior convergence under the IID setting. However, it's noteworthy that the Scaffold loss being consistently zero raises concerns and calls for further investigation to ensure the integrity of the experimental setup or to identify possible anomalies in the algorithm's behavior.

\subsection{Non-IID Experimental Results}
\begin{figure}[!htb]
\centering
\begin{subfigure}{\linewidth}
\includegraphics[width=\linewidth]{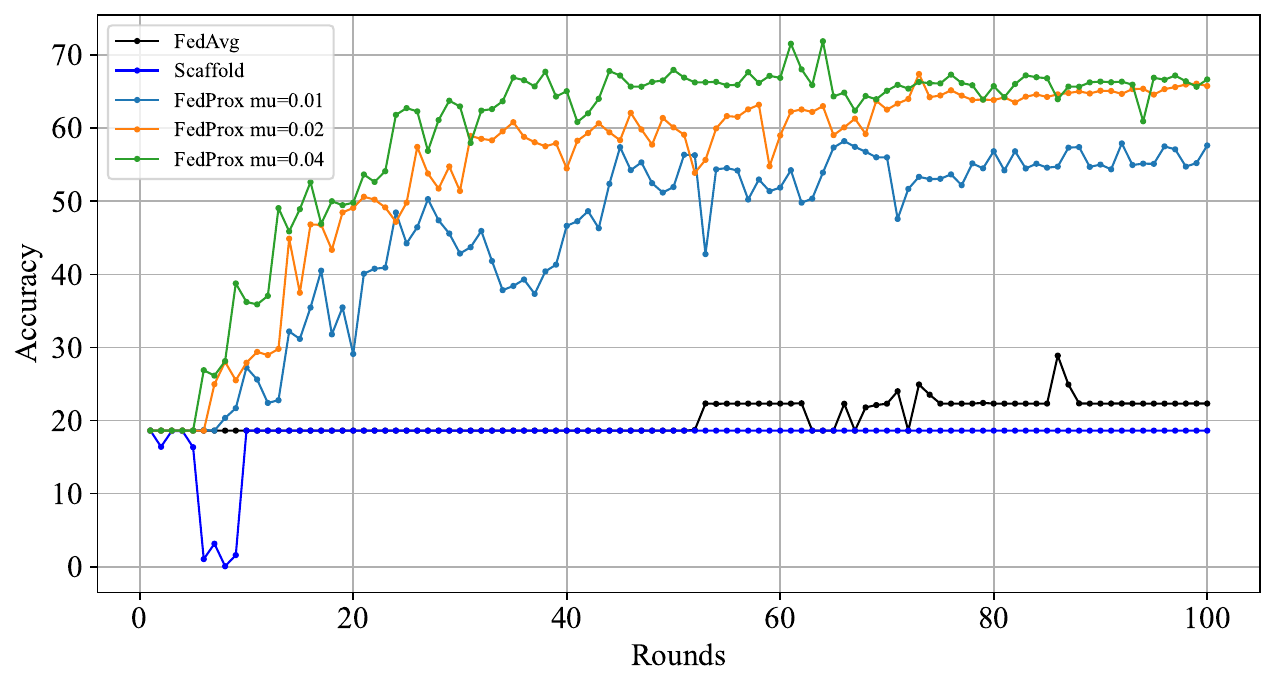}
\caption{Comparison of accuracy of the collaborative learning under non-IID considerations with existing methods.}
\label{fig:noniid-global-acc}
\end{subfigure}
\begin{subfigure}{\linewidth}
\includegraphics[width=\linewidth]{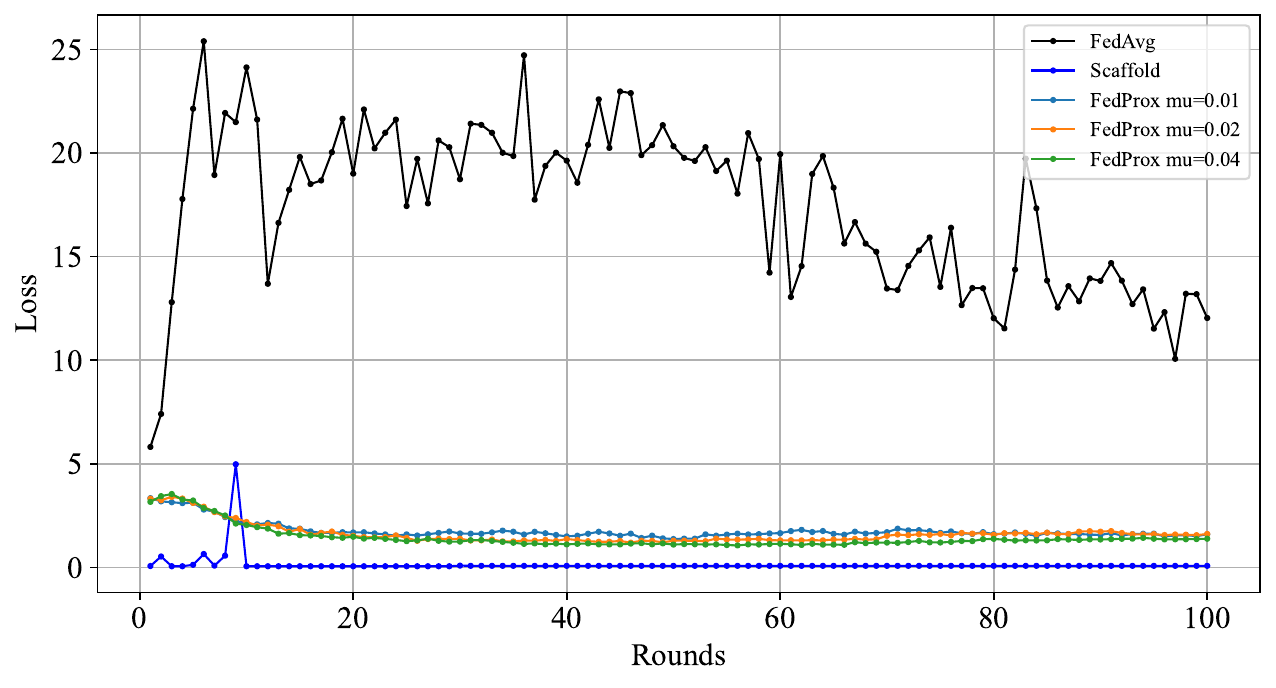}
\caption{Comparison of loss incurred in collaborative learning under non-IID considerations with existing methods.}
\label{fig:noniid-global-loss}
\end{subfigure}
\caption{Global performance outcomes of the proposal under non-IID data in contrast with conventional methods.}
\label{fig:noniid-global}
\end{figure}

In the non-IID experimental setting detailed in Fig.~\ref{fig:noniid-global}, the performance metrics of the evaluated FL algorithms' distinct behaviors. For the FedAvg algorithm, the global accuracy remains consistently low at 28.88\%, indicating challenges in adapting to the non-IID data distribution. The global loss exhibits fluctuating patterns, peaking at 12.54. Contrastingly, the FedProx algorithm with mu=0.01 displays an initial period of marginal improvement in accuracy, reaching 58.46\%, while the loss demonstrates a declining trend, settling at 1.58. However, setting mu to 0.04 results in a more substantial improvement, with the accuracy peaking at 71.88\% and the loss decreasing to 1.10. Notably, the Scaffold maintains stable performance, with an accuracy of 18.78\% and minimal loss values. These results indicate that FedProx, especially with mu = 0.04, shows enhanced adaptability and convergence in the face of non-IID data distributions compared to the FedAvg and Scaffold algorithms.

\section{Conclusion}
\label{sec:conclusion}
In conclusion, the comparative study of FL algorithms for IoT attack detection reveals insights into their performance under different data distribution settings. In the IID, FedAvg struggles with non-uniform data distributions, resulting in lower accuracy, whereas FedProx demonstrates improved adaptability to non-IID settings, exhibiting better accuracy and loss metrics. The scaffold maintains stable performance across varying data distributions but requires further optimization for enhanced effectiveness.
However, notable limitations must be addressed. In non-IID settings, handling statistical heterogeneity remains challenging, potentially leading to suboptimal global model performance. Scalability concerns also arise with increasing numbers of clients, impacting coordination mechanisms and overall efficiency.
Future research could explore advanced FL algorithms to handle statistical heterogeneity in detecting IoT attacks. Developing lightweight models could improve scalability and reduce computational overhead. Assessing robustness against adversarial attacks is crucial for FL system security, and benchmarking against established intrusion detection systems would offer insights into efficacy and areas for improvement. These efforts are vital for advancing the practical applicability and robustness of federated learning in IoT attack detection.

\bibliographystyle{IEEEtran}
\bibliography{refs.bib}

\end{document}